\documentclass[letterpaper, 10 pt, conference]{ieeeconf}  

\IEEEoverridecommandlockouts                              

\usepackage{balance} 
\usepackage{amsmath,amsfonts}
\usepackage{algorithm,algpseudocode}
\usepackage{graphicx}
\usepackage{textcomp}
\usepackage[utf8]{inputenc}
\usepackage{multirow}
\usepackage{units}
\usepackage{tabularx}
\usepackage{paralist}
\usepackage{booktabs} 
\usepackage{hyperref}

\title{\LARGE \bf
Decentralized Multi-Agent Reinforcement Learning\\ with Global State Prediction
}

\author{Joshua Bloom, Pranjal Paliwal, Apratim Mukherjee, and Carlo Pinciroli%
\thanks{All authors are with Robotics Engineering, Worcester Polytechnic Institute, Worcester, MA, USA (email: {\tt\small \{jdbloom, ppaliwal, amukherjee, cpinciroli\}@wpi.edu}.}}

\begin{document}
\maketitle
\thispagestyle{empty}
\pagestyle{empty}

\begin{abstract}
Deep reinforcement learning (DRL) has seen remarkable success in the control of single robots. However, applying DRL to robot swarms presents significant challenges. A critical challenge is non-stationarity, which occurs when two or more robots update individual or shared policies concurrently, thereby engaging in an interdependent training process with no guarantees of convergence. Circumventing non-stationarity typically involves training the robots with global information about other agents' states and/or actions. In contrast, in this paper we explore how to remove the need for global information. We pose our problem as a Partially Observable Markov Decision Process, due to the absence of global knowledge on other agents. Using collective transport as a testbed scenario, we study two approaches to multi-agent training. In the first, the robots exchange no messages, and are trained to rely on implicit communication through push-and-pull on the object to transport. In the second approach, we introduce Global State Prediction (GSP), a network trained to form a belief over the swarm as a whole and predict its future states. We provide a comprehensive study over four well-known deep reinforcement learning algorithms in environments with obstacles, measuring performance as the successful transport of the object to a goal location within a desired time-frame. Through an ablation study, we show that including GSP boosts performance and increases robustness when compared with methods that use global knowledge. 

\end{abstract}


\section{Introduction}


Reinforcement learning (RL) has shown promise in the control of single agents. However, coordinating multi-agent systems through RL (MARL) still poses significant challenges. Among them, a critical one is \textit{non-stationarity}, which occurs when independently learning agents change their policy during training. This breaks the Markov assumption, which in turn removes convergence guarantees due to the likelihood of agents entering an infinite loop of mutual adaptation around sub-optimal policies.

Non-stationarity is typically circumvented through diverse techniques, which include the use of a centralized policy controlling all agents simultaneously, introducing global state or action information during training, or performing complex message passing \cite{gronauer_multi-agent_2022,amato_modeling_2019,foerster_learning_2016,wu_spatial_2021,papoudakis_dealing_2019}.

In this paper, we explore how to circumvent non-stationarity without resorting to the agents having explicit knowledge of each other's actions and states. We analyze two approaches: one in which the agents do not exchange messages, and one in which the agents exchange minimal, partial, and local information and learn to build a prediction of the global state.

We showcase our approach in a collective transport scenario, in which a team of robots is physically connected around an object that must reach a predefined location. This scenario is representative of the larger category of \textit{robotic aggregates}, i.e., teams of robots connected to each other to form a rigid lattice. Modular robots \cite{sun_salamanderbot_2020}, self-assembling robots \cite{mondada_swarm-bot_2004}, swarm formation control \cite{barnes_unmanned_2007}, and certain approaches to collective transport \cite{tuci_cooperative_2018} fall into this category. Controlling aggregates is in itself an open research question for MARL due to additional challenges such as continuous control and partial observability of the global state.

In our experimental scenario, implicit (i.e., message-free) communication (IC) occurs through the natural push-and-pull that the robots experience by moving the object. We show that this form of communication is sufficient for the robots to learn a control policy capable of emergent coordination.

In a second set of experiments, we introduce Global State Prediction (GSP). GSP is a neural network that collects partial local observations and predicts future state changes at the team level. This prediction is then fed as an additional input to the control policy network. GSP removes the need for explicit global information, while imposing minimal communication requirements.

To train these approaches, we propose Aggregate Centralized Training with Decentralized Execution (A-CTDE). Through this technique, robots store their local experiences in a shared memory for single policy training. This policy is then placed on all robots for decentralized execution. Unlike Centralized Training with Decentralized Execution, all stored local experiences have been directly influenced by the other agents and not simply treated as part of the environment. 

We compare IC and GSP against a previously proposed method where robots share Global Knowledge (GK) via direct broadcast communication \cite{zhang_decentralized_2020}. We study the performance of four well-known training methods: DQN, DDQN, DDPG, and TD3. Our results indicate that robots operating only with IC are able to complete the task with relatively high success. When GSP is added, our results out-perform IC and even GK. GSP has lower communication requirements than GK, and also eliminates the need for global knowledge. We present an in-depth analysis on the behaviors across all three methods and explore the effect each robots' shared information has on GSP.


\section{Background and Related Work}


\begin{figure*}
  \centering
  \includegraphics[width=.9\textwidth]{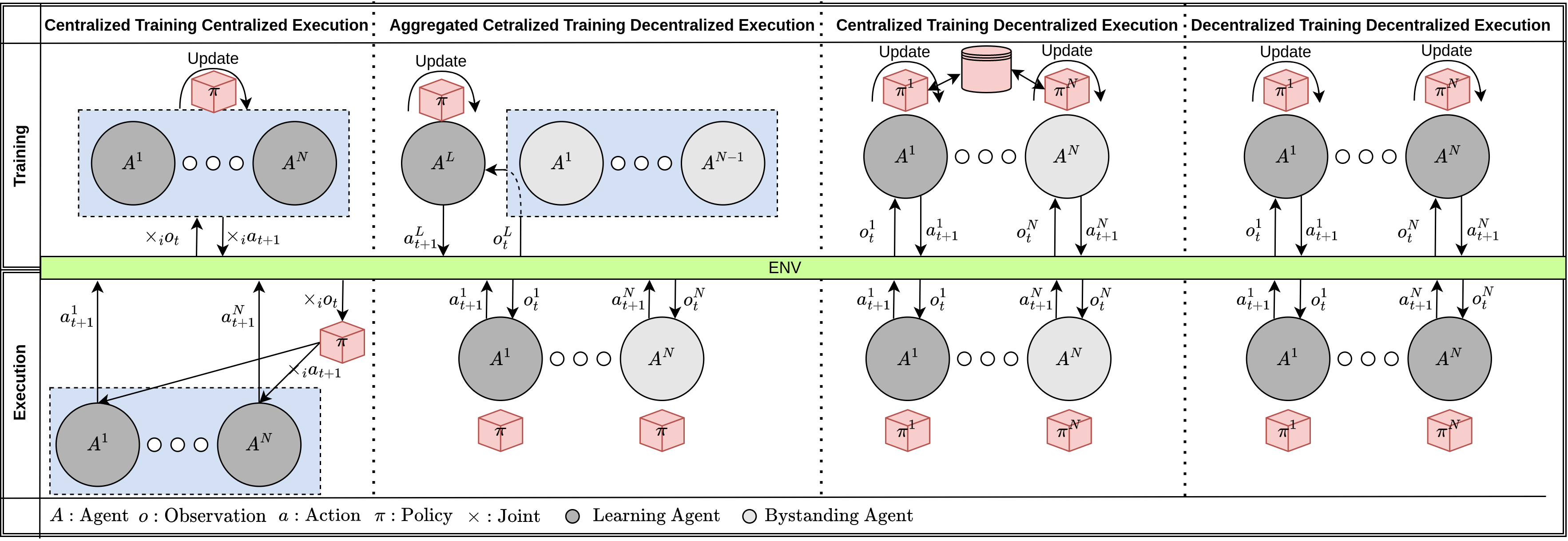}
  \caption{Common schemes for training and executing Deep Reinforcement Learning Algorithms for Multi-Agent environments. CTCE trains a single policy on global information and produces a joint action. A-CTDE trains a single policy on local information and single agent actions, is then copied and placed on each actor for local execution. CTDE trains multiple policies on local information with the addition of global information, the global information is removed during execution. DTDE trains agent policies independently on local information and executes independently on local information}
  \label{fig:training_schemes}
\end{figure*}

\subsection{Deep Reinforcement Learning}
Reinforcement Learning (RL) is typically formalized as a  Markov Decision Process $\langle S, A, R, T\rangle$, where $S$ is the set of all possible discrete states, $A$ is the set of all available discrete actions, $R(s, a, s'):S\times A\times S\rightarrow \mathbb{R}$ is a reward function, and $T(s'|s,a)$ is a probabilistic transition function mapping states and actions to new states. Optimal decision-making in an MQP is captured by the Bellman Equation:
\begin{equation*}
    Q(s, a) = \sum_{s'}T(s'|s, a)[R(s, a, s') + \gamma \max_{a'} Q(s', a')],
\end{equation*}
where $Q(s, a)$ is a function mapping state-action pairs to a reward, and $\gamma$ is a discount factor.

RL has shown success when states and actions are discrete, and when the number of state-action pairs is limited. However, when working with robots, it is often unavoidable to consider continuous state spaces. Deep Reinforcement Learning (DRL) offers a solution by expressing the mapping $Q(s,a)$ as a function $Q(s, a|\theta)$ where $\theta$ is a set of neural network parameters.

Using DRL in realistic environments often presents the issue of noisy or incomplete observations. This is typically captured as a Partially Observable Markov Decision Process $\langle S, A, R, T, \Omega, O\rangle$, where $o\in\Omega(s)$ is a partial observation of the full state $s\in S$ according to a probabilistic function $O(o|s)$.

Here we employ four well known DRL algorithms: Deep Q Network (DQN) \cite{mnih_human-level_2015}, Double DQN (DDQN) \cite{van_hasselt_deep_2016}, Deep Deterministic Policy Gradient (DDPG) \cite{lillicrap_continuous_2015}, and Twin Delayed DDPG (TD3) \cite{fujimoto2018addressing}. We chose these four DRL implementations due to their recent advancement in current robotics research \cite{morales_survey_2021}. 

Note the difference in control. DQN and DDQN approximate the $Q(s,a)$ and, as such, return expected future values for each possible action given a state. An $\epsilon$-greedy training schema is then employed where, with probability $(1-\epsilon)$, the agent chooses greedily over the action space, and with probability $\epsilon$ it chooses randomly over the action space. Parameter $\epsilon$ is decayed until reaching a minimum set value. DQN and DDQN require a discrete action space. Alternatively, DDPG and TD3 directly output the actions to take, or the \textit{policy}, as a real-valued number. Therefore, the action space is continuous. 

\subsection{Multi-Agent Reinforcement Learning}
Moving from a single agent to a multi-agent setting introduces several complexities in learning \cite{hernandez-leal_survey_2019}. There are various approaches to tackling multi-agent scenarios, typically categorized in terms of distributedness across two axes: training and execution \cite{gronauer_multi-agent_2022}. As shown in Figure \ref{fig:training_schemes}(A), Centralized Training with Centralized Execution (CTCE) employs a single trained policy that observes a global joint observation space and selects a joint action for all agents to execute. CTCE is a natural extension for DRL methods to multi-agent scenarios. However, the state-action space grows exponentially with the number of agents. 

In contrast, Decentralized Training with Decentralized Execution (DTDE), shown in Figure \ref{fig:training_schemes}(D) trains every agent independently, treating all other agents as part of the environment. Ideally, this learning schema is desirable because it theoretically splits a large, global training problem into more manageable local instances. However, this approach does not guarantee convergence because of non-stationarity, which breaks the Markov assumption. 

To cope with non-stationarity, one can combine the benefits of a centralized trainer with the flexibility of decentralized execution (CTDE), as shown in Figure \ref{fig:training_schemes}(C). In this approach, each agent trains its own local policy with the addition of external global information only available during training. This external information is then removed during execution. Foerster \textit{et al.} \cite{foerster_learning_2016} introduce learners who communicate (CTDE), allowing for differentiation, via discretized/regularized units (DRU), between agents during training. DRUs provide a way to directly calculate the gradient corresponding to how well other agents received messages. While CTDE methods like MADDPG \cite{lowe_multi-agent_2020}, A3C \cite{mnih_asynchronous_2016}, MATD3 \cite{ackermann_reducing_2019}, PPO \cite{yu_surprising_2022}, VDN \cite{sunehag_value-decomposition_2017}, and Q-Mix \cite{rashid_qmix_2018} have shown remarkable success in MARL when they know the positions of all agents involved, they fail to learn when the information shared is restricted to just the local information \cite{nayak_scalable_2023}. Our work seeks to leverage the benefits of developing a single policy shared amongst all of the agents, such as resilience to attrition and repeatability given identical observations while operating on local information only. 

A recent, alternative idea for dealing with non-stationarity in DTDE or CTDE is to locally model individual agents and produce a belief over their future actions \cite{papoudakis_dealing_2019, rabinowitz_machine_2018}. Wu \textit{et. al} \cite{wu_spatial_2021} introduce \textit{spatial intention maps} which combine with the agent's observation space to provide spatial information about other agent's actions. Agents in this environment communicate their belief over other agents' actions via a set of waypoints indicating predicted trajectories. The agents compile these communicated trajectories and build a global map of beliefs. Other work incorporating belief in the multi-agent setting includes reasoning directly what your own policy would have you do given another agents' observations \cite{raileanu_modeling_2018, stulp_implicit_2006}, and rewarding causal influence over other agents' actions \cite{jaques_social_2019,ndousse_emergent_2021}.

\subsection{Our Contribution}
While powerful, the mentioned methods based on belief are limited in their prediction abilities, because reasoning over agents individually is prone to scaling problems. In this work, we propose a new approach to modeling belief over agents in a MARL setting called Global State Prediction (GSP). GSP is rooted in the human ability to psychologically represent others, often referred to as \textit{Theory of Mind} \cite{premack_does_1978}. However, rather than forming a belief over each agent individually, GSP abstracts further, similar to \textit{herd mentality} \cite{raafat_herding_2009}. GSP is specifically trained to predict the outcome of all agents' actions given a limited amount of local partial information from each agent. Unlike auxiliary tasks \cite{jaderberg_reinforcement_2016, hessel_multi-task_2018, shelhamer_loss_2017, mirowski_learning_2017, baisero_learning_2020, pathak_curiosity-driven_2017}, GSP is a separate network whose output is used only as an observation by each agents' policy to aid with action selection.


\section{Methodology}

\begin{figure}
  \centering
  \includegraphics[width=75mm]{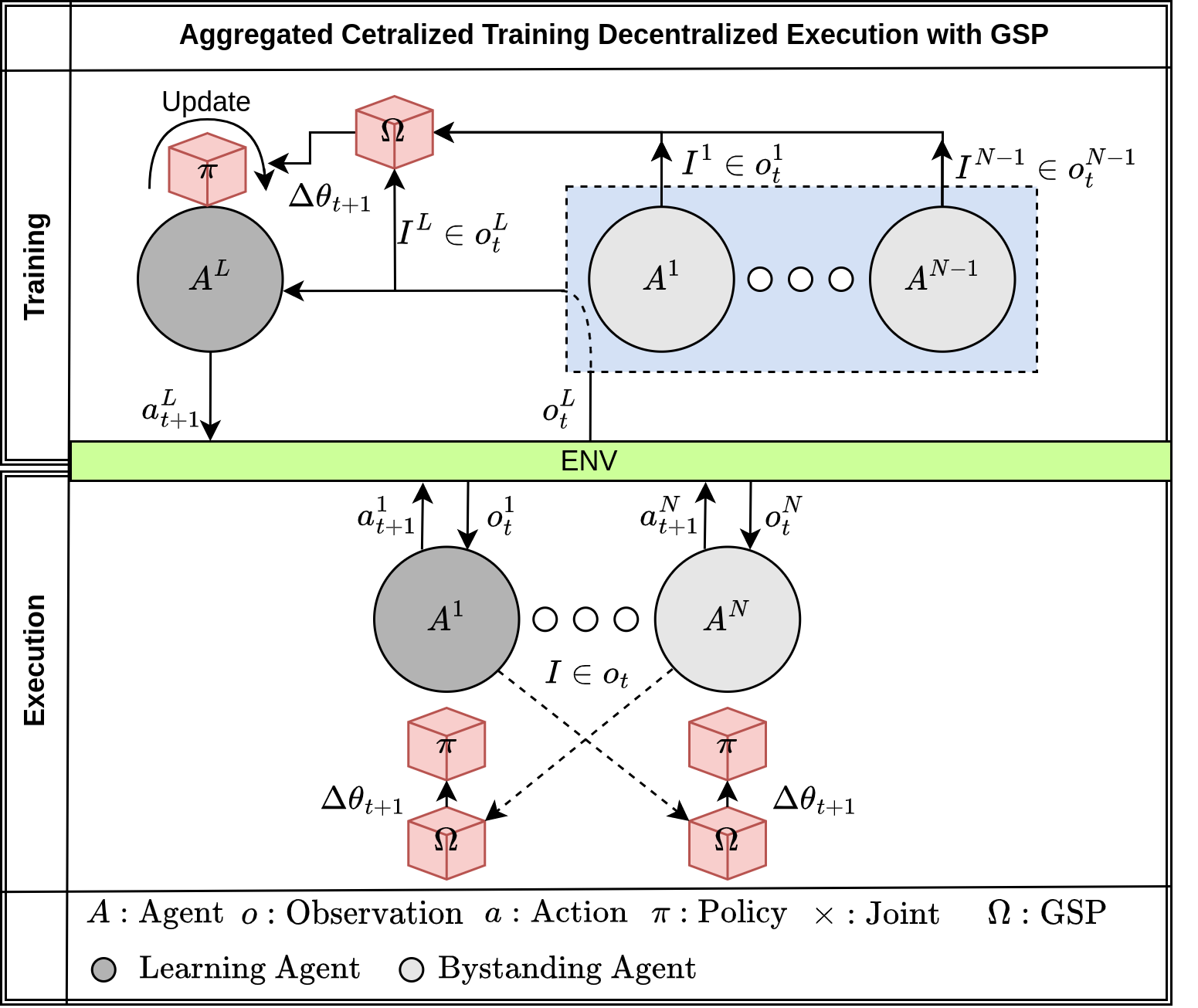}
  \caption{GSP implementation with A-CTDE}
  \label{fig:A-CTDE-GSP}
\end{figure}

GSP works by collecting a subspace of each robots' observation space $\vec I \in O$ producing a prediction on the global state of the robot aggregate in the future. Unlike Zhang \textit{et. al} \cite{zhang_decentralized_2020}, our method does not require full state space information. Rather, it is sufficient to share aggregate information, e.g., the average of each robots' proximity readings. This input is both less informative and more realistic to retrieve and communicate than global knowledge as defined in \cite{zhang_decentralized_2020}. GSP then takes these averaged readings and produces a prediction over the change in orientation of the robot aggregate. This prediction is appended to the local observation spaces of each robot for action selection as shown in Figure \ref{fig:A-CTDE-GSP}.

\subsection{Aggregate Centralized Training Decentralized Execution}
We implement a version of Centralized Training with Decentralized Execution (CTDE) similar to Radulescu \textit{et al.} \cite{atzmueller_deep_2019}, however, our method differs due to the physical connections between learning agents. This connection allows agents to mutually affect each others' states after action execution. In this implementation, all of the agents store their local experiences in a shared replay, from which a single policy is learned. This allows us to bootstrap the policy, and because we are learning from local experiences without global knowledge, this local policy can be scaled after training to more or less agents than was trained. We will refer to this version as Aggregate Centralized Training Decentralized Execution (A-CTDE), as shown in Figure \ref{fig:training_schemes}(B). 

During training, everything is centralized and the policy is queried once per agent per time-step, taking as input the local observation of that agent and producing an action for that agent to take. Once all actions are compiled, the agents execute their respective actions. At execution, this single policy is copied and placed on each agent individually for that agent to run locally and asynchronously.

Although this method of training allows us to treat training as a single agent, it is still susceptible to environmental non-stationarity because, even though all the agents are running the same policy, they are concurrently updating this shared policy \cite{atzmueller_deep_2019}. From the perspective of a single agent, all of the other agent's policies are changing through time. 

\subsection{Addressing Environmental Non-Stationarity in Robot Aggregates}
Environmental non-stationarity of a single agent $i$ can be modeled in the transition probability function $T(s_i'|s_i, a_i, \rho_{\pi(\theta)})$, where $\rho_{\pi(\theta)}$ is the transition dynamics following a hidden and changing distribution defined by the current parameters $\theta$ of the policy $\pi$. 

This hidden and changing distribution $\rho$ is the direct cause of environmental non-stationarity. If there was only one learning agent present in the environment, then $\rho$ would be hidden but not changing and, as such, would be learnable. But, because all of the agents are updating their shared policy concurrently, the hidden distribution defining $\rho$ is also moving, thus leading to the moving target problem \cite{gronauer_multi-agent_2022}.

We explore dealing with environmental non-stationarity in three different ways: implicit communication (IC), globally shared knowledge (GK), and global state prediction (GSP).

\subsubsection{Implicit Communication}
By the definition of a robot aggregate in this paper, all of the robots are rigidly attached in some pre-defined lattice, naturally inducing implicit communication via push-and-pull on their connection to the other robots. This push-and-pull is observed by all of the other agents in the error they experience when they execute a selected action. As shown in Sec. \ref{sec:evaluation}, this form of implicit communication proves to be sufficient for DRL to adequately model $\rho_{\pi(\theta)}$ and produce viable policies. 

\subsubsection{Global Knowledge}
As a baseline for our work, we consider the use of complete and global knowledge (GK) on the robot actions and states. We implement the method presented by Zhang \textit{et al.} \cite{zhang_decentralized_2020}, in which the robots broadcast the current global position and observed velocity vectors $(\vec x, \vec{\dot x})$ of every robot in the aggregate. The set of all position and observed velocity vectors excluding the observing robot's vectors is $(X^{-i}, \dot X^{-i})$ and is appended to the local observation spaces of every robot $i$. By observing $(X^{-i}, \dot X^{-i})$ the changing effect $\rho$ exhibits as a result of concurrent updates on $\pi$ is eliminated because the robots are communicating their policy $\pi(s)$ allowing robot $i$ to understand the joint action space $\times A_t$. Thus, non-stationarity is eliminated, allowing DRL to reach viable policies. 

\subsubsection{Global State Prediction}
Communicating $(\vec x, \vec{\dot x})$ is not always possible or accurate, and may not construe all relevant information to the environment, such as observations from sensors. Furthermore, global information about the robot aggregate, such as positions, may not be available in most realistic scenarios. In keeping with the example presented in \cite{zhang_decentralized_2020}, GSP receives only $\dot X^{-i}$ and provides a prediction over the change in orientation of the robot aggregate. Due to the centralized training present in A-CTDE, the actual change in angle is known in the following time-step for GSP to be trained on. The prediction of the change in angle is a prediction of the result over the joint action $\times A_t$ taken during the current time-step thus eliminating the moving target problem present in $\rho$ and allowing DRL to construct viable policies.



\section{Evaluation}
\label{sec:evaluation}


\begin{figure}
  \centering
  \includegraphics[width=75mm]{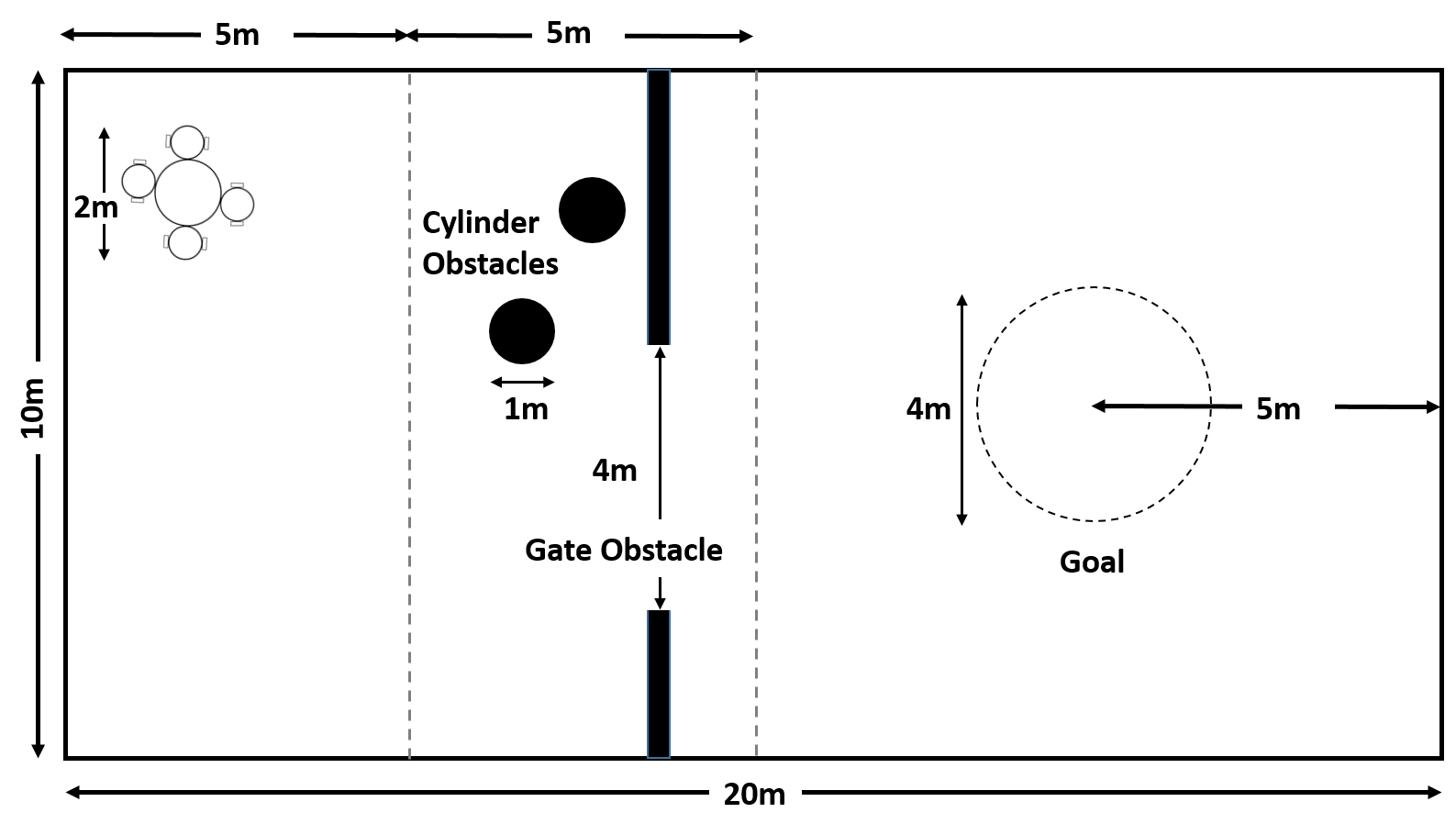}
  \caption{Evaluation environment for a collective transport task with obstacles present. The left-most region represents the robot aggregate generation zone, the center region represents the obstacle generation zone, and the region on the right represents the goal}
  \label{fig:environment_setup}
\end{figure}

\begin{figure}
  \centering
  \includegraphics[width=75mm]{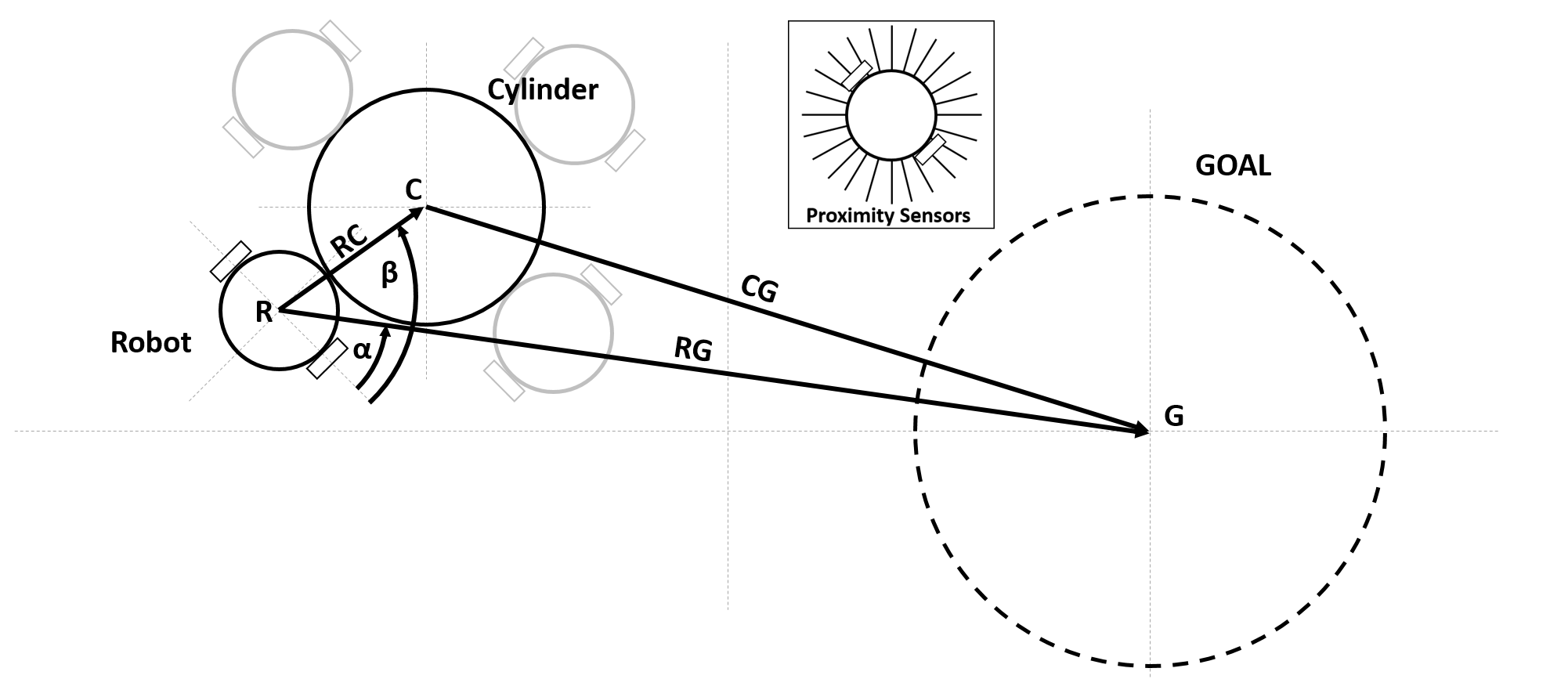}
  \caption{Robot observation space consisting of the vectors from the robot to the object to transport and to the goal, the distance from the object to transport to the goal, the wheel velocities, and the array of 24 proximity sensors}
  \label{fig:robot_observations}
\end{figure}

\subsection{Experimental Setup}
\subsubsection{Collective Transport}
We evaluate all methods, IC, GK, and GSP, on a collective transport task as shown in Figure \ref{fig:environment_setup}. This task is simulated using the ARGoS multi-robot simulator \cite{pinciroli_argos_2012} with the Buzz programming language \cite{pinciroli_buzz_2016} for individual robot control and the PyTorch library for the Neural Network infrastructure.

The environment is defined such that the robot aggregate will be generated in the left-most region, where the pose of the object to transport is randomly generated according to a uniform distribution. The robots are then randomly placed at equal mutual distances around the object. 

The central region of the environment is the \textit{obstacle generation zone}. We evaluate our work against two types of obstacles: cylinder obstacles and a gate obstacle. The positions of the cylinder obstacles are randomly generated according to a uniform distribution. The horizontal position of the gate obstacle and the vertical position of the center of the opening are randomly generated according to a uniform distribution. Both the cylinder obstacles and the gate obstacle are constrained to be within the obstacle generation zone which lies between the \textit{robot aggregate generation zone} and the \textit{goal zone}. 

\subsubsection{Robot Control and Sensing}
We chose the foot-bot \cite{bonani_marxbot_2010} due to its differential-drive controller and independent non-actuated turret attached to an actuated gripper. The turret allows the robot to rotate independently from the gripper. The gripper is not actively controlled and is only actuated upon initialization or failure. 

The robots are controllable through wheel velocities $\unit[v\in(-10, 10)]{cm/s}$. The agent chooses a $\unit[\Delta v \in \{-0.1, 0, 0.1\}]{cm/s}$ in the case of DQN and DDQN, and $\Delta v\in[-0.1, 0.1]\unit{cm/s}$ in the case of DDPG and TD3. Note the distinction between a discrete action space and a continuous action space as DQN and DDQN choose their actions via an epsilon-greedy strategy, whereas DDPG and TD3 output their actions directly. In the case of DQN and DDQN, we must provide an action space consisting of all possible combinations of actions between the two wheels, represented as $|\Delta v|^2$. We choose a discretization of $|\Delta v|=3$ for control simplicity.

All robots in the environment are identical and are able to sense several values as shown in Figure \ref{fig:robot_observations}. They observe the vector from themselves to the goal $\vec{RG}$, the vector from themselves to the object to transport $\vec{RC}$, the distance from the object to transport and the goal $|\vec{CG}|$, their wheel speeds $\textbf{v}$, and their 24 proximity sensors $\mathbf{p} $ uniformly distributed around the robot. The proximity sensors can sense a distance of \unit[2]{m} returning a value of 0 for no observation and 1 for touching the sensed object. 

\subsubsection{Global State Prediction} GSP only requires the robots to communicate their average proximity values $\sum \textbf{p} / |\textbf{p}|$, a single floating-point value. These values are used as input to GSP, which then produces a predicted change in the orientation of the robot aggregate as a result of the actions executed at the current time-step. This prediction is appended to the observation space of the robots. GSP collects raw sensory information as opposed to GK presented in Zhang \textit{et al.} \cite{zhang_decentralized_2020} which uses actual speeds and positions without noise.

\begin{algorithm}[t]
\caption{IC}\label{alg:actde-IC}
\begin{footnotesize}
\begin{algorithmic}
\State Initialize Policy $\pi_0$ and Buffer $B$
\For {\textbf{each} episode}
\State Initialize Robots, Object, and Obstacles
\State Receive initial observations $O_t$
\While {\textbf{not} done at timestep $t$}
\For {\textbf{each} robot $i$}
\State $a^i_t \leftarrow \pi_t(o^i_t)$
\EndFor
\State Execute actions $A_t$ and receive ($O_{t+1}$, $R_t$, Done)
\For {\textbf{each} robot $i$}
\State  $B \leftarrow (o_t^i$, $a_t^i$, $r_t^i$, $o_{t+1}^i$, Done)
\EndFor
\State $\pi_{t+1}$ $\leftarrow$ Update Policy $\pi_t$
\State $o_t \leftarrow o_{t+1}$
\EndWhile
\EndFor
\end{algorithmic}
\end{footnotesize}
\end{algorithm}

\subsubsection{Training}
We train policies in environments with either two cylinder obstacles or a gate obstacle. When training with a gate obstacle, we employ curriculum learning \cite{bengio_curriculum_2009}, with the opening distance of the gate starting equal to the vertical width of the environment (i.e, no obstacle). The opening is then shortened by \unit[0.5]{m} at a fixed episodic frequency until the opening distance reaches the desired minimum.

\begin{table*}[t]
  \begin{center}
    \caption{Success Rates with Obstacles}
    \label{tab:Obstacles}
    \begin{tabular}{c|c|c|c|c|c|c|c|c}
      \toprule 
       Obstacles &DQN-IC &DQN-GSP &DDQN-IC &DDQN-GSP &DDPG-IC &DDPG-GSP &TD3-IC &TD3-GSP\\
      \midrule 
       2  &73\% &86\% &90\% &94\% &91\% &94\% &91\% &97\%\\
       4  &56\% &84\% &81\% &82\% &86\% &87\% &82\% &85\%\\
       Gate &72\% &73\% &74\% &80\% &77\% &80\% &76\% &84\%\\
      \bottomrule 
    \end{tabular}
  \end{center}
\end{table*}

The reward function used for all training rewards moving in the direction of the goal while penalizing proximity readings and time taken to complete the task:
\begin{equation}
    R_{i, t}=-2+\frac{\vec{CG}\cdot\bigg(C(x_t, y_t)-C(x_{t-1}, y_{t-1})\bigg)}{\bigg|\vec{CG}\bigg|\cdot\bigg|C(x_t, y_t)-C(x_{t-1}, y_{t-1})\bigg|} - \frac{1}{|\textbf{p}|}\sum\textbf{p}.
\end{equation}
Algorithm \ref{alg:actde-IC} reports how A-CTDE works with IC. Due to the physical connections in the robot aggregate, we are able to train a single policy, trained on local observations, that is executed on every robot. 

\begin{figure}
  \centering
  \includegraphics[width=75mm]{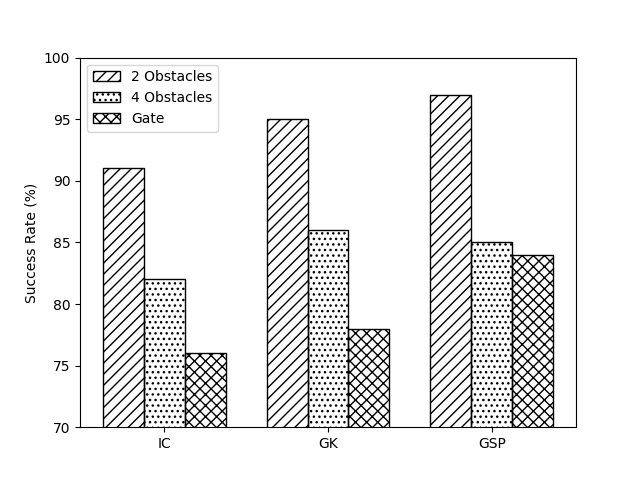}
  \caption{Comparison of TD3 using Implicit Communication (IC), Global Knowledge (GK), and Global State Prediction (GSP) evaluated on 2 obstacles, 4 obstacles, and the gate obstacle}
  \label{fig:GK_comparison}
\end{figure}

We formalize GSP in Algorithm \ref{alg:actde-GSP}. Here, all of the averaged proximity reading $\textbf{P}$ are broadcasted and used as input to the GSP network $\Omega(\textbf{P}) \rightarrow \Delta \theta$, where $\Delta\theta$ is the predicted change in the orientation of the robot aggregate. During training, this prediction is evaluated in the next time-step against the ground truth.

\subsubsection{Neural Network Architectures and Hyperparameters}
All networks architectures are built using three fully connected layers activated using ReLU and optimized using ADAM. DQN and DDQN have size (31, 64, 128, 9) and DDPG and TD3 have size (31, 400, 300, 2). We train GSP using DDPG.

For DQN and DDQN, we drew inspiration from Mnih \textit{et al.} \cite{mnih_human-level_2015} for our choice in hyperparameters. Specifically, a discount factor $\gamma$ of 0.99997 and a learning rate $\eta$ of $10^{-4}$ were used. We used mini-batch learning with a memory of $10^6$ experiences and a batch size of 100, where an experience is comprised of the initial state, action taken, reward gained, the resultant state, and a boolean terminal flag. We used an $\epsilon$-greedy policy with a linear decay of $10^{-6}$ and a minimum $\epsilon$ value of 0.01. The target network was updated every 1,000 learning iterations and, during training, learning occurred every time-step. For DDPG and TD3 we set hyperparameter values identical to Lillicrap \textit{et al.} \cite{lillicrap_continuous_2015}. GSP was trained via DDPG using the same network architecture and hyperparameters.

\subsection{Experimental Evaluation}

\begin{algorithm}
\caption{GSP}\label{alg:actde-GSP}
\begin{footnotesize}
\begin{algorithmic}
\State Initialize Policy $\pi_0$ and Buffer $B_1$
\State Initialize GSP $\Omega$ and Buffer $B_2$
\For {\textbf{each} episode $t$}
\State Initialize Robots, Object, and Obstacles
\State Receive initial observations $O_t$
\For {\textbf{each} robot $i$}
\State Broadcast $\textbf{P}_t^i = \sum \textbf{p}^i_t / \textbf{p}^i_t$
\EndFor
\State $\Delta \theta_t \leftarrow \Omega(\textbf{P}_t)$ 
\While {\textbf{not} done at timestep $t$}
\For {\textbf{each} robot i}
\State $a^i_t \leftarrow \pi_t(o^i_t, \Delta \theta_t)$
\EndFor
\State Execute actions $A_t$ and receive ($O_{t+1}$, $R_t, R^{\Delta \theta}_t$, Done)
\For {\textbf{each} robot $i$}
\State Broadcast $\textbf{P}_{t+1}^i = \frac{1}{\textbf{p}^i_{t+1}}\sum \textbf{p}^i_{t+1}$
\EndFor
\State $\Delta \theta_{t+1} \leftarrow \Omega(\textbf{P}_{t+1})$ 
\For {\textbf{each} robot $i$}
\State  $B_1 \leftarrow ((o_t^i, \Delta \theta_t$), $a_t^i$, $r_t^i$, ($o_{t+1}^i, \Delta \theta_{t+1}$), Done)
\EndFor
\State $B_2 \leftarrow (\textbf{P}_t, \Delta t \theta_t, R_t^{\Delta \theta}, \textbf{P}_{t+1}$, Done)
\State Update Policy $\pi$ and GSP $\Omega$
\State $o_t \leftarrow o_{t+1}$, $\Delta \theta_t \leftarrow \Delta \theta_{t+1}$
\EndWhile
\EndFor
\end{algorithmic}
\end{footnotesize}
\end{algorithm}

\subsubsection{Obstacles}
We experiment on 4-robot aggregates in three obstacle environments: 2-cylinder obstacles, 4-cylinder obstacles, and the gate obstacle. To avoid evaluating on training environments, the random seed was changed from training and kept to the same value across all experiments. By keeping the seed the same across testing experiments, we are able to directly compare policies on the same environments. 

We compare the successful transport of the object to the goal location within the required time. Each policy was tested in 100 different random orientations of robots, object to transport and obstacles. We report the percentage of successful runs in Table \ref{tab:Obstacles}.

Obstacles provide an interesting evaluation scenario because they provide stimulus to some robots in the aggregate, but not all robots may be able to observe them. Thus, a more sophisticated degree of coordination is required over an environment with no obstacles, where the robots may learn to simply drive in the direction of the goal. We train and test on the 2-cylinder obstacle environment as a baseline and then increase the complexity to 4-cylinder obstacles.

\begin{figure}
  \centering
  \includegraphics[width=80mm]{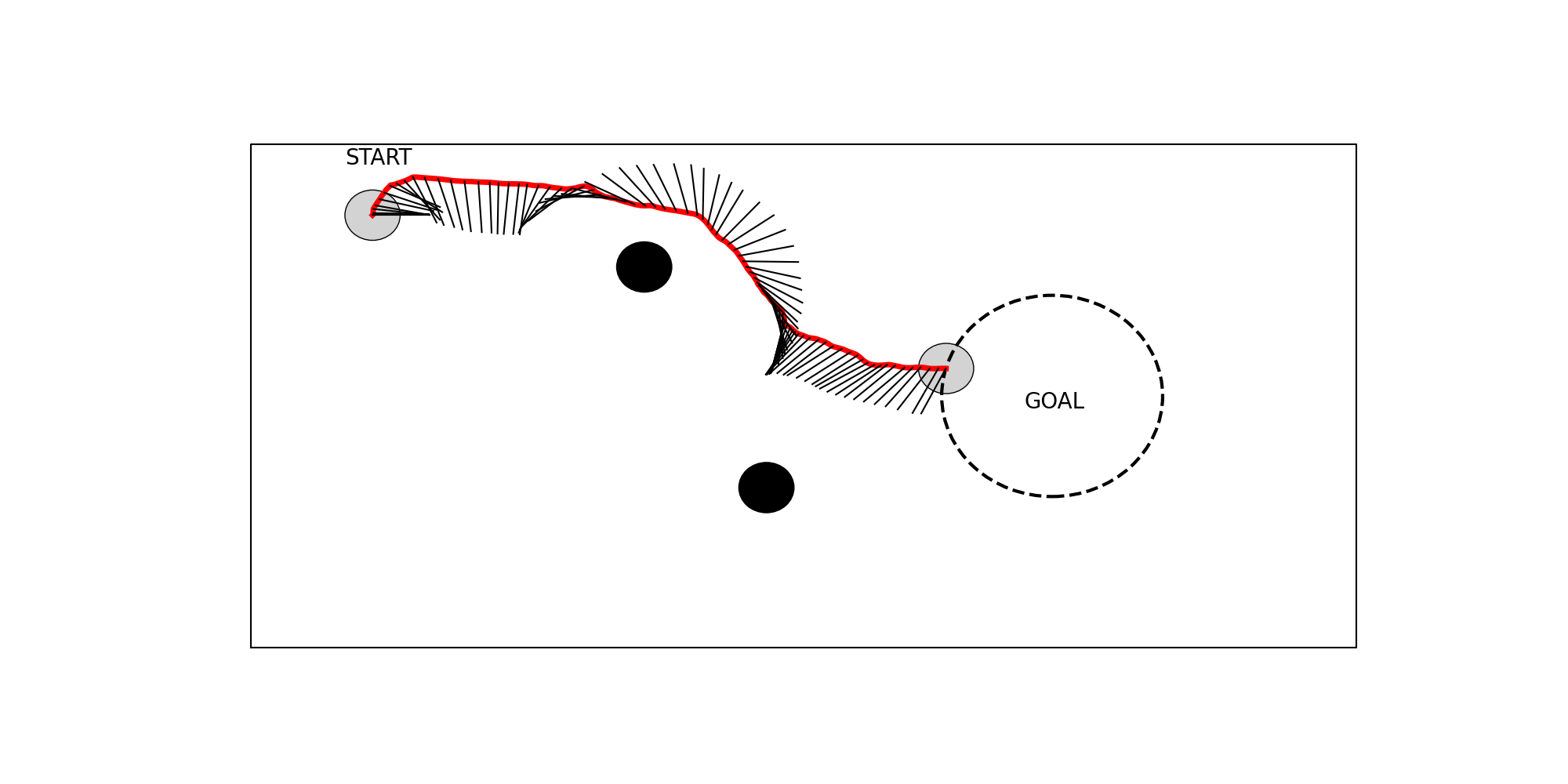}
  \caption{Trajectory Behavior of DQN with GSP plotted with the orientation of the robot aggregate over the episode}
  \label{fig:Obstacle_Avoidance_GSP}
\end{figure}

The results allow us to draw two conclusions. Firstly, as the complexity of the learning algorithm increases, so does the success rate. This result is in line with previous findings in the literature. Secondly, GSP increases the success rate on all obstacles for all methods. In the case of DQN, the increase is substantial, with success rates on the 4-obstacle environment increasing by 28\%. In addition, when we observe the behavior of DRL with GSP, as shown in Figure \ref{fig:Obstacle_Avoidance_GSP}, we notice active obstacle avoidance, depicted by the changing orientation of the robot aggregate shown as the black lines.

\begin{figure}
  \centering
  \includegraphics[width=80mm]{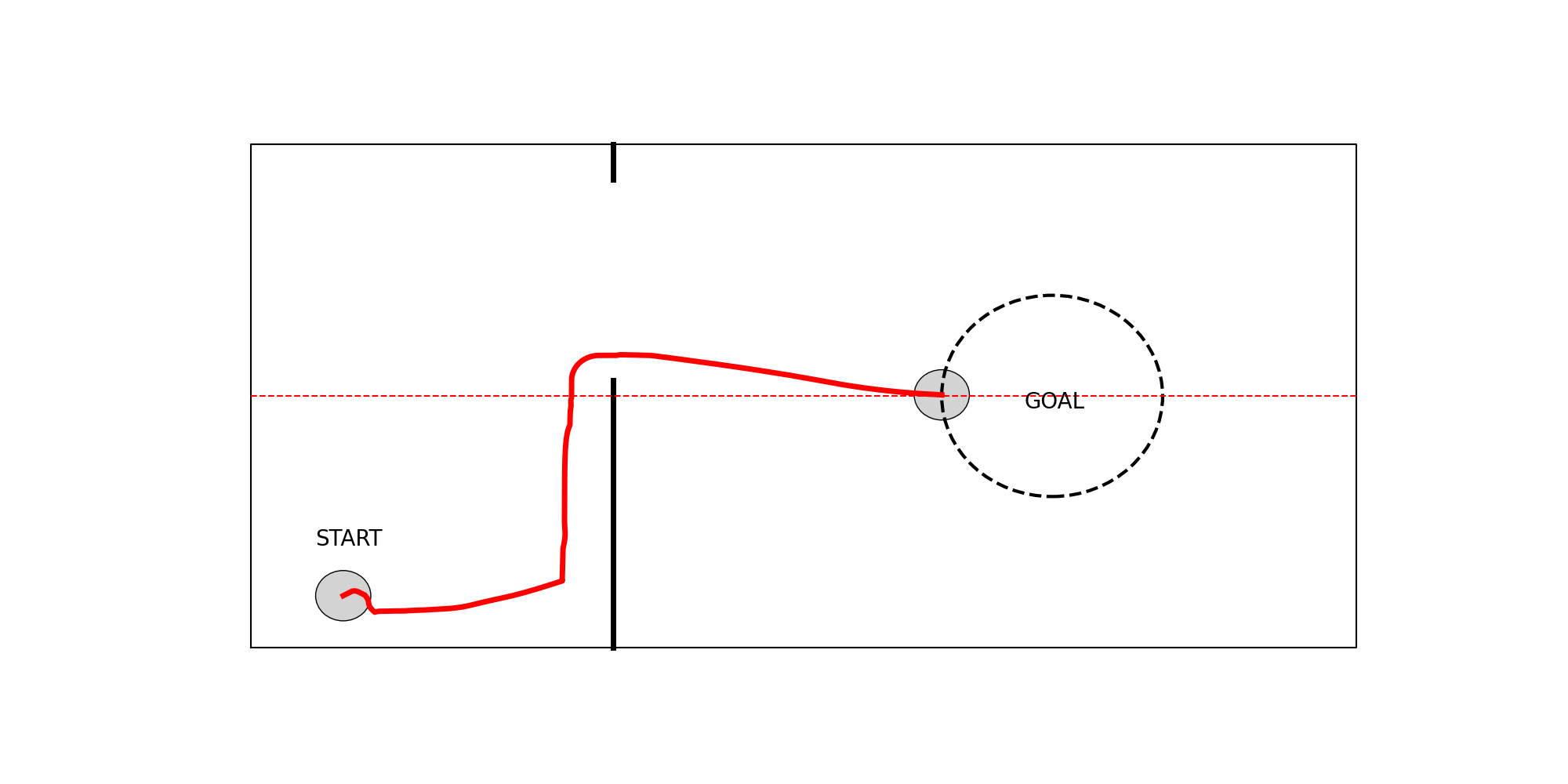}
  \caption{Trajectory Behavior of DQN with GSP on the gate obstacle with the opening on the opposite side of the environment from the initialization of the robot aggregate}
  \label{fig:gate_success_gsp}
\end{figure}

The gate obstacle proves to be harder than the 2-obstacle environment. A variant of the bug algorithms is a reasonable behavior for this type of obstacle, where the robot aggregate follows the wall until it finds the opening. However, IC is unable to learn it. We attribute this inability to the need for the robot aggregate to occasionally move \textit{away} from the target to reach the opening in certain environments. However, as shown in Figure \ref{fig:gate_success_gsp}, GSP learns a successful strategy.

When we evaluate IC and GSP against GK, we observe that the success rates increases in the 2-obstacle environment and the gate environment when using GSP over using GK. The 4-obstacle environment shows similar results. GK performed similarly across all methods, Figure \ref{fig:GK_comparison} reports the results for TD3. This is an interesting result, considering GSP uses less information than GK, and also forgoes the need for global knowledge. 

\subsubsection{Resilience}
Resilience is an important feature for multi-robot systems. For robot aggregates, in particular, failures introduce asymmetries in the dynamics. We define a failure as a complete loss of power resulting in a disengagement between the robot and the aggregate. We explore an 8-robot aggregate with a chance of having up to 75\% of the robots fail during an episode. During episode initialization, we select the number of robots to fail by giving each robot a 25\% chance of failure up to the desired 75\% of robot failures. Once the number of failures is determined we randomly assign a failure time.

Table \ref{tab:Resillience} reports our findings evaluating IC, GK, and GSP on TD3 in 2- and 4-cylinder obstacle environments. In both environments GSP outperforms GK and substantially increases the success rate over IC.  

\subsubsection{GSP Network Analysis}
\begin{figure}
  \centering
  \includegraphics[width=80mm]{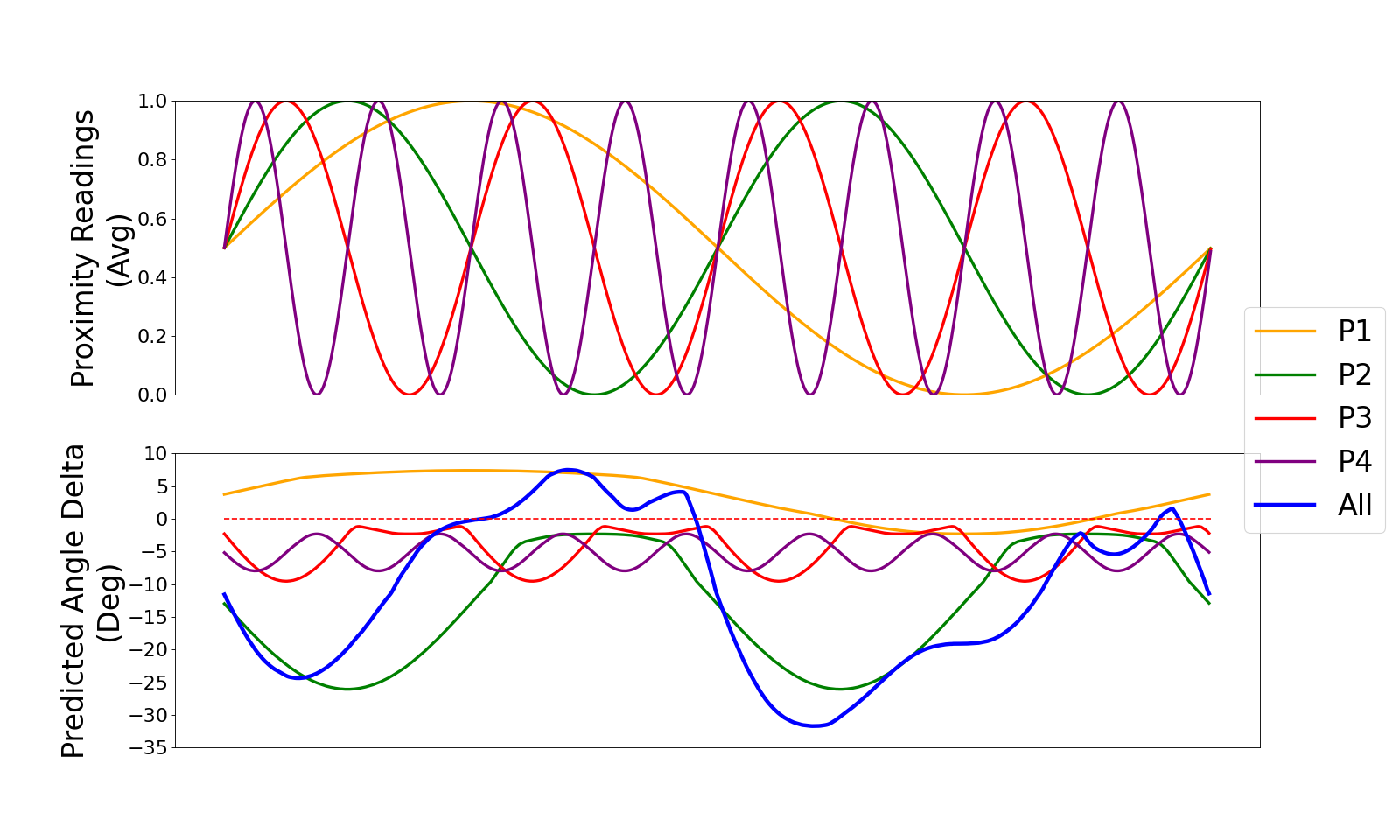}
  \caption{Top: Simulated averaged proximity values for a 4 robot aggregate; Bottom: GSP network predicted change in angle corresponding to single robot inputs and all inputs combined}
  \label{fig:GSP_probe}
\end{figure}

To gain insight into the behavior of GSP, we present an analysis of a trained GSP network in a 4-robot aggregate, shown in Figure \ref{fig:GSP_probe}. We simulate the robot aggregate's proximity readings \{P1, \dots, P4\} communicated across the robots with sine functions over a period of $2\pi$ with the frequency increasing for each robot. By increasing the frequency, we can better visualize the additional value that input has on the predicted change in orientation angle. 

The bottom plot shows the predicted change in orientation angle, where each line is the result of only receiving the corresponding sine function in the top plot. For example, the P1 line shown in orange at the bottom is the result of only receiving the P1 sine wave and keeping other inputs 0. 

\begin{table}[t]
  \begin{center}
    \caption{Success Rates with Obstacles and Failures}
    \label{tab:Resillience}
    \begin{tabular}{c|c|c|c}
      \toprule 
       Obstacles &TD3-IC &TD3-GK &TD3-GSP\\
      \midrule 
       2  &83\% &89\% &95\%\\
       4  &67\% &79\% &81\%\\
      \bottomrule 
    \end{tabular}
  \end{center}
\end{table}

\begin{figure}
  \centering
  \includegraphics[width=80mm]{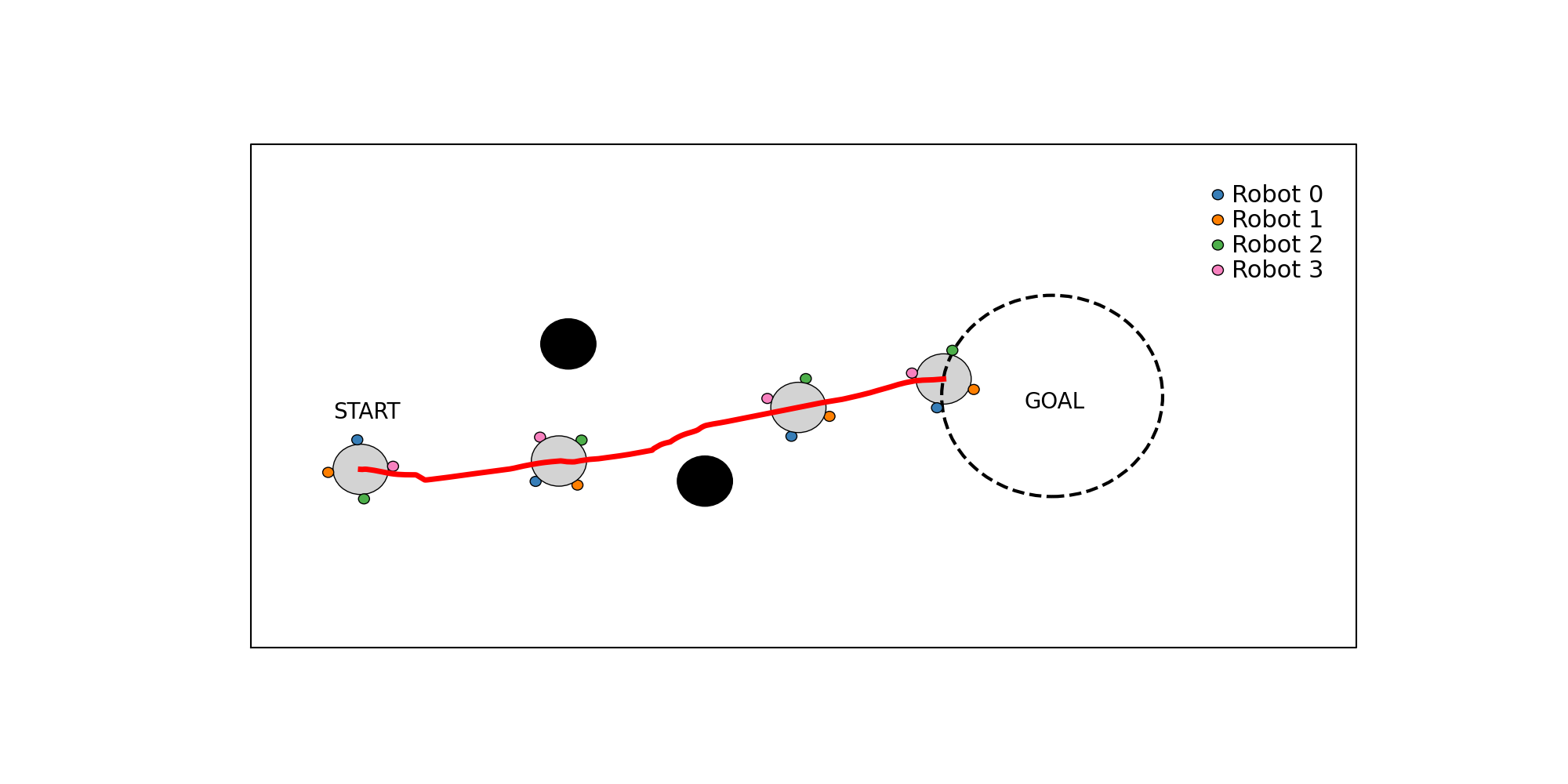}
  \caption{Emergent reorientation behavior of the robot aggregate with GSP when solving a 2-cylinder obstacle environment}
  \label{fig:Robot_Orientations}
\end{figure}

From this analysis, we notice that all robots play a role in predicting the change in orientation angle of the robot aggregate. However, some robots have a larger impact than others. Robots 0 and 1 are key to generating a larger prediction, with robot 2 playing a moderate role and robot 3 having the smallest impact on the prediction. Upon further investigation, as shown in Figure \ref{fig:Robot_Orientations}, we found that the robot aggregate reorients at the start of an episode so that robot 1 faces the goal, robots 0 and 2 faces the north and south walls, respectively, and robot 3 is on the backside, opposite of robot 1. This form of role allocation is an emergent property that was not explicitly rewarded during training.



\section{Conclusion}

We present two approaches to addressing non-stationarity in MARL without the use of global information, evaluated in a collective transport scenario. The first approach is based on implicit communication realized through pushing and pulling the object instead of message passing. The second approach consists of endowing the robots with a network trained to predict the future state of the system (in our case the object) by aggregating partial local observations. We evaluate the performance of our method through four well known reinforcement learning algorithms.

We show that IC is sufficient to achieve coordinated collective transport in environments with obstacles. Furthermore, we demonstrate the ability of GSP to perform better than a prior method that used global knowledge. We provide an in-depth analysis into the mechanisms driving coordination. 

We plan on further developing the idea of minimizing non-stationarity through global state prediction from partial observations. In particular, we will apply this concept to tasks beyond collective transport, such as collective motion and multi-robot foraging, and will investigate how more complex MARL methods such as MADDPG \cite{lowe_multi-agent_2020} and A3C \cite{hernandez-leal_agent_2019} perform along side GSP.

\bibliographystyle{IEEEtran} 
\bibliography{IEEEabrv, references}

\end{document}